\documentclass[sigconf]{acmart}

\AtBeginDocument{%
  }

\setcopyright{acmlicensed}
\copyrightyear{2025}
\acmYear{2025}
\acmDOI{XXXXXXX.XXXXXXX}

\acmConference[CIKM '25]{Proceedings of the 2025 ACM Conference on Information and Knowledge Management}{October 21--25, 2025}{Birmingham, UK}
\acmISBN{978-1-4503-XXXX-X/25/10}

\usepackage{graphicx}
\usepackage{subcaption}
\usepackage{amsmath}
\usepackage{booktabs}
\usepackage{url}
\usepackage{xcolor}
\usepackage{tikz}
\usetikzlibrary{arrows.meta, positioning, shapes.geometric}

\begin{document}

\title{Guess the Age of Photos: An Interactive Web Platform for Historical Image Age Estimation}

\author{Hasan Yucedag}
\authornote{Supervised by Adam Jatowt.}
\email{hasan.yuecedag@student.uibk.ac.at}
\affiliation{%
  \institution{Leopold-Franzens Universit\"{a}t Innsbruck}
  \city{Innsbruck}
  \country{Austria}}

\author{Adam Jatowt}
\email{adam.jatowt@uibk.ac.at}
\affiliation{%
  \institution{Leopold-Franzens Universit\"{a}t Innsbruck}
  \city{Innsbruck}
  \country{Austria}}

\renewcommand{\shortauthors}{Y\"{u}ceda\u{g} and Jatowt}

\begin{abstract}
This paper introduces \textit{Guess the Age of Photos}, a web platform engaging users in estimating the years of historical photographs through two gamified modes: \textit{Guess the Year} (predicting a single image’s year) and \textit{Timeline Challenge} (comparing two images to identify the older). Built with Python, Flask, Bootstrap, and PostgreSQL, it uses a 10,150-image subset of the Date Estimation in the Wild dataset (1930--1999). Features like dynamic scoring and leaderboards boost engagement. Evaluated with 113 users and 15,473 gameplays, the platform earned a 4.25/5 satisfaction rating. Users excelled in relative comparisons (65.9\% accuracy) over absolute year guesses (25.6\% accuracy), with older decades easier to identify. The platform serves as an educational tool, fostering historical awareness and analytical skills via interactive exploration of visual heritage. Furthermore, the platform provides a valuable resource for studying human perception of temporal cues in images and could be used to generate annotated data for training and evaluating computer vision models.
\end{abstract}




\maketitle

\section{Motivation}
Historical photographs are invaluable artifacts, providing deep insights into societal, cultural, and technological evolution across decades. They preserve moments such as World War II battlefields, the evolution of 1950s fashion, or the architectural shifts in urban landscapes, serving as tangible connections to the past~\cite{Lowenthal_1998}. However, estimating their age poses significant challenges for non-experts, requiring specialized knowledge of visual cues like clothing styles, photographic techniques (e.g., black-and-white vs. color film), or historical events. This knowledge gap limits public engagement with these resources, particularly in educational settings where interactive learning is increasingly valued.

Gamification has emerged as a powerful strategy to enhance engagement and learning by transforming tasks into interactive, rewarding experiences~\cite{deterding2011gamification, nicholson2015gamification}. Despite the abundance of online historical image archives (e.g., Library of Congress, Europeana), few platforms leverage gamification to make the exploration of these images both entertaining and educational. Existing tools, such as static image databases or basic quizzes, lack the dynamic feedback and competitive elements that drive participation~\cite{smith2006uses}.

\textit{Guess the Age of Photos} addresses this gap by creating a gamified web application with two modes: \textit{Guess the Year} and \textit{Timeline Challenge}. Utilizing a curated 10,150-image dataset and features like dynamic scoring, leaderboards, and instant feedback, it fosters historical awareness and analytical skills among diverse users~\cite{seaborn2015gamification}. Targeting students, educators, history enthusiasts, and casual users, it makes heritage exploration both accessible and engaging, contributing to cultural education and public appreciation of historical narratives~\cite{Lowenthal_1998}. Beyond entertainment and heritage appreciation, this platform provides a unique environment for studying human perception of temporal cues in visual data. The gameplay data, reflecting user strategies and biases in age estimation, can be analyzed to understand how humans process and interpret visual information related to time. Furthermore, the platform can be used to generate a large, annotated dataset of images with human-estimated ages, valuable for training and evaluating computer vision models in the task of automatic image age estimation.

\section{Related Work}
The \textit{Guess the Age of Photos} platform builds on research in gamification, image age estimation, and heritage education. Gamification enhances user engagement in educational contexts by integrating game design elements like points and leaderboards, which studies show increase motivation and retention~\cite{deterding2011gamification, hamari2014gamification, nicholson2015gamification, seaborn2015gamification}. In educational settings, gamified platforms have demonstrated improved learning outcomes by making complex subjects enjoyable and interactive~\cite{dominguez2013gamifying}.

Automatic image age estimation has progressed with datasets like Date Estimation in the Wild (DEW), enabling machine learning models—particularly convolutional neural networks—to achieve high accuracy by analyzing visual features such as color histograms or edge patterns~\cite{Muller2017, springstein2020deep}. M\"{u}ller et al. showcased robust results on Flickr images~\cite{Muller2017}, while subsequent work refined temporal classification through advanced feature extraction~\cite{fernando2020temporal}. Human perception, by contrast, relies on intuitive cues, offering less precision but richer contextual insights~\cite{tversky1974judgment, rhodes2009age}.

Heritage education advocates for interactive methods to propagate historical knowledge, emphasizing digital platforms that make cultural heritage accessible to broader audiences~\cite{Lowenthal_1998, smith2006uses, hooper2012heritage}. Our platform aligns with this trend, combining gamification with human and computational approaches to enhance educational impact and public engagement with history.

\section{Dataset Description}
The platform leverages a 10,150-image subset of the DEW dataset~\cite{Muller2017}, covering the period from 1930 to 1999 with approximately 145 images per year. Sourced from Flickr, these images come with metadata including \texttt{img\_id}, \texttt{gt\_year}, \texttt{date\_taken}, \texttt{date\_granularity}, and \texttt{url}, stored in \texttt{meta.csv}. The images were retrieved using the DEW-Downloader tool~\cite{DEWRepo} and resized to 800x600 pixels using Python’s Pillow library to optimize gameplay performance. Stored in \texttt{/home/user/finalimages/}, the dataset ensures reliability despite potential URL inconsistencies. The variable \texttt{date\_granularity} was addressed by assigning representative years, with manual validation to maintain accuracy, resulting in a diverse and authentic collection for historical exploration.

\section{System Building Description}
\textit{Guess the Age of Photos} is a client-server web application developed using Python, Flask, Bootstrap, and PostgreSQL, hosted on the university server at \texttt{disc-imageguessing.uibk.ac.at}\footnote{Access the live demo at \url{http://disc-imageguessing.uibk.ac.at/login}~\cite{demo2025}.}. The Flask backend handles game logic, user authentication with secure password hashing, and data management through routes such as \texttt{/login}, \texttt{/guess\_the\_year}, and \texttt{/timeline\_challenge}. A PostgreSQL database organizes user profiles, image metadata, and gameplay logs across tables like \texttt{users}, \texttt{images}, and \texttt{game\_plays}, with encrypted user data to ensure privacy. Images are served from a dedicated server directory, with Gunicorn and Nginx providing scalability through load balancing and caching mechanisms~\cite{DEWRepo}.

\begin{figure}[t]
  \centering
  \begin{tikzpicture}[
      scale=0.75,
      transform shape,
      node distance=0.6cm and 0.6cm,
      auto,
      box/.style={
        rectangle,
        draw,
        rounded corners,
        minimum height=0.8cm,
        minimum width=1.4cm,
        align=center,
        fill=blue!10,
        thick
      },
      db/.style={
        rectangle,
        draw,
        rounded corners,
        minimum height=0.8cm,
        minimum width=1.2cm,
        align=center,
        fill=green!10,
        thick
      },
      storage/.style={
        rectangle,
        draw,
        rounded corners,
        minimum height=0.8cm,
        minimum width=1.2cm,
        align=center,
        fill=yellow!10,
        thick
      },
      arrow/.style={
        ->,
        thick,
        black
      },
      label/.style={
        font=\tiny,
        midway,
        fill=white,
        inner sep=0.3pt
      }
    ]
    \node[box] (client) {User\\(Browser)};
    \node[box, right=of client] (nginx) {Nginx};
    \node[box, right=of nginx] (gunicorn) {Gunicorn};
    \node[box, right=of gunicorn] (flask) {Flask\\Backend};
    \node[db, below=of flask] (db) {PostgreSQL\\DB};
    \node[storage, below=of gunicorn] (storage) {Image\\Repo};

    \draw[arrow] (client) -- (nginx) node[label, above] {Requests};
    \draw[arrow] (nginx) -- (gunicorn);
    \draw[arrow] (gunicorn) -- (flask);
    \draw[arrow] (flask) -- (db) node[label, right] {Fetch/Store};
    \draw[arrow] (db) -- (flask);
    \draw[arrow] (flask) -- (storage) node[label, above left] {Load};
    \draw[arrow] (storage) -- (flask);
    \draw[arrow] (flask) -- (gunicorn);
    \draw[arrow] (gunicorn) -- (nginx);
    \draw[arrow] (nginx) -- (client) node[label, below] {Responses};
    
  \end{tikzpicture}
  \caption{System architecture of \textit{Guess the Age of Photos}, showing request and response flow.}
  \Description{Diagram illustrating the flow of user requests from a web browser through Nginx and Gunicorn to the Flask backend, which retrieves data from a PostgreSQL database and images from a repository, returning responses to the user.}
  \label{fig:architecture}
\end{figure}
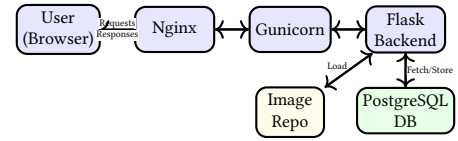

The Bootstrap frontend delivers responsive HTML templates rendered via Jinja2, enhanced with JavaScript and AJAX for real-time score updates, image refreshes, and interactive feedback (e.g., displaying titles of the images after a guess). Demo mode allows unregistered access for casual play, while registered users benefit from personalized score tracking, leaderboards, and performance analytics. Random image selection with hidden ground truth years ensures fairness and replayability. Figure~\ref{fig:architecture} illustrates the architecture, detailing the flow from user requests through Nginx and Gunicorn to the Flask backend, interacting with PostgreSQL and the image repository.

\subsection{Game Modes}
\begin{figure}[t]
  \centering
  \begin{subfigure}{0.48\textwidth}
      \centering
      \includegraphics[width=\textwidth]{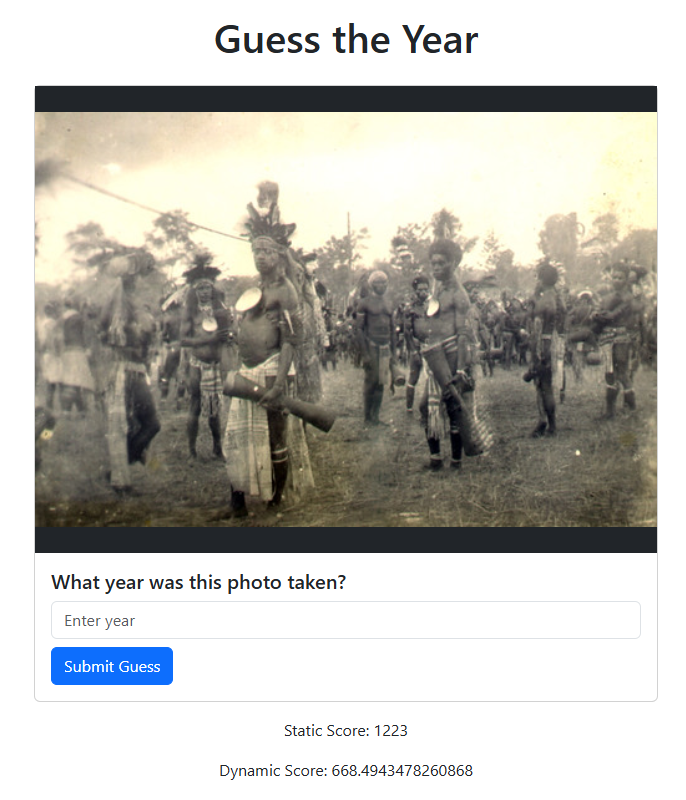}
      \caption{Guess the Year mode. This image was taken in 1944.}
      \Description{A historical image centered on the screen with a numeric input field below for entering a year guess (1930--1999).}
  \end{subfigure}
  \hfill
  \begin{subfigure}{0.48\textwidth}
      \centering
      \includegraphics[width=\textwidth]{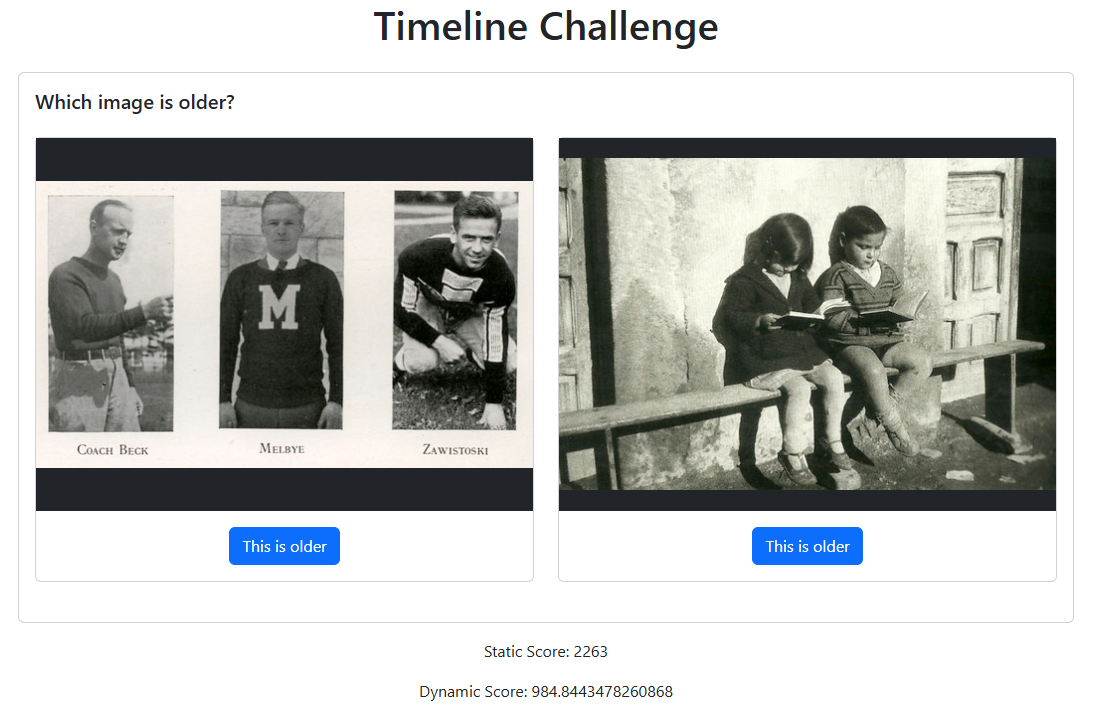}
      \caption{Timeline Challenge mode. Users choose the older image between the one from 1933 (left) and 1930 (right).}
      \Description{Two historical images side by side, each with a button below labeled “Left is older” or “Right is older” for selection.}
  \end{subfigure}
  \caption{Interfaces of the two game modes in \textit{Guess the Age of Photos}.}
  \label{fig:game_modes}
\end{figure}

The platform offers two game modes (see Figure~\ref{fig:game_modes}):
\begin{itemize}
    \item \textbf{Guess the Year}: Users estimate a single image’s year (1930--1999). A static score of 10 points is awarded for guesses within $\pm5$ years, with dynamic points calculated as $10 \times \left(1 - \frac{\min(|\text{guess} - \text{actual}|, 100)}{100}\right)$. Feedback includes the correct year and title of the image.
    \item \textbf{Timeline Challenge}: Users compare two images to select the older. Correct answers earn 10 points, with dynamic bonuses based on the year gap $\Delta$: $5.0 \times \begin{cases} 1.0 & \text{if } \Delta \leq 10 \\ 1.0 - \frac{\Delta - 10}{40} & \text{if } 10 < \Delta \leq 50 \\ 0.1 & \text{if } \Delta > 50 \end{cases}$. Feedback displays both years, points, and comparative insights (e.g., “Left image is from year x and the Right image is from year y")~\cite{tversky1974judgment}.
\end{itemize}

\section{Evaluation}
From January 20 to March 30, 2025, the platform was actively used by 113 users, resulting in a total of 15,473 gameplays (2,252 \textit{Guess the Year}, 13,221 \textit{Timeline Challenge}), indicating strong engagement. All participating users were based in Austria. A significant portion came from the high school BORG Innsbruck, around a quarter were students at the University of Innsbruck, and the rest included friends, family, and acquaintances of the developer.

\begin{figure}[t]
  \centering
  \includegraphics[width=0.6\textwidth]{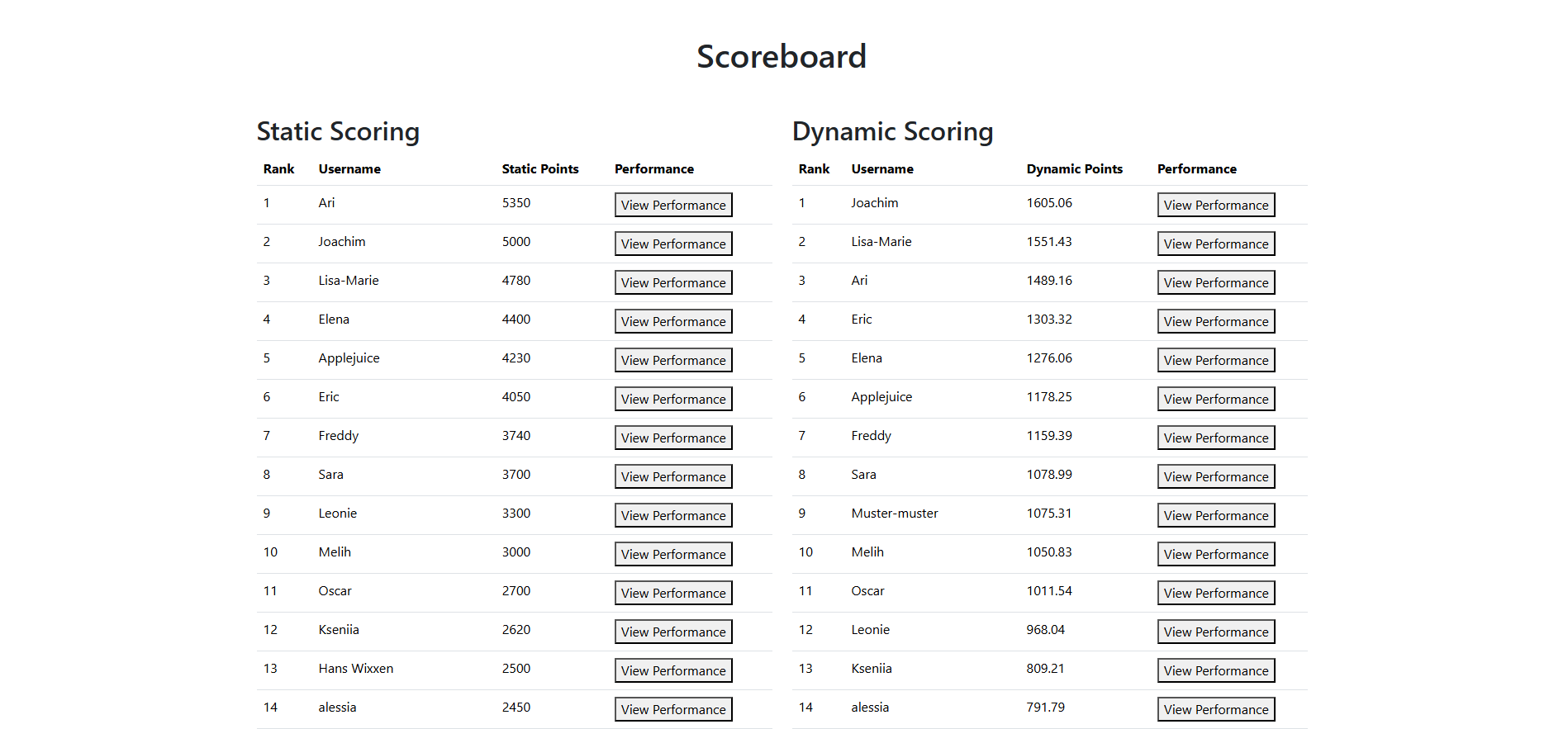}
  \caption{Leaderboard interface displaying user rankings.}
  \Description{A table with two sections: "Static Scoring" and "Dynamic Scoring." Each section has columns for Rank, Username, Points (Static or Dynamic), and a "View Performance" button. Rows list users (e.g., Ari, Joachim, Lisa-Marie) with scores (e.g., 5350 static points for Ari, 1605.06 dynamic points for Joachim).}
  \label{fig:leaderboard}
\end{figure}

The leaderboard (Figure~\ref{fig:leaderboard}) highlights the competitive element, ranking users based on both static and dynamic point systems~\cite{hamari2014gamification}. Performance data (Figure~\ref{fig:performance_analysis}) shows variations in accuracy across different decades, with higher accuracy for older images (e.g., 1930s–1950s) and a decline for more recent ones.

\subsection{User Feedback}
Feedback from 20--24 users, representing diverse educational backgrounds (high school to postgraduate), showed high satisfaction: 91.3\% rated enjoyment 1 or 2 (1 = very good, 5 = very bad), 95.8\% found mechanics easy to understand, and 79.2\% rated navigation very easy. The overall rating averaged 4.25/5, reflecting usability and appeal~\cite{seaborn2015gamification}. Users appreciated instant feedback and leaderboard visibility but suggested challenges like distinguishing 1970s and 1980s styles, prompting ideas for adaptive difficulty or hint systems~\cite{dominguez2013gamifying}.

\subsection{Accuracy Analysis}
\begin{figure}[t]
  \centering
  \includegraphics[width=0.6\textwidth]{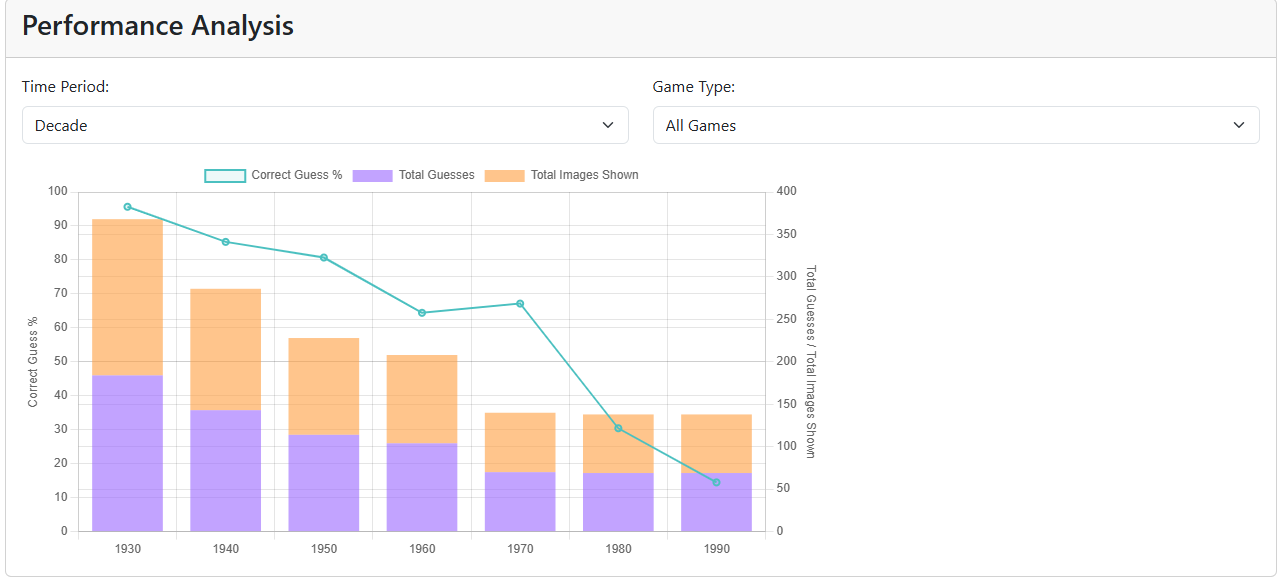}
  \caption{Performance analysis across decades for both game modes combined.}
  \Description{A chart showing performance metrics per decade (1930s--1990s). Stacked bars in purple (total guesses) and orange (total images shown) are on the left y-axis (0--400). A teal line (correct guess percentage) is on the right y-axis (0--100), peaking at ~90\% in the 1930s and declining to ~20\% in the 1990s.}
  \label{fig:performance_analysis}
\end{figure}

\textit{Timeline Challenge} achieved 65.9\% accuracy (8,708/13,221 correct), outperforming \textit{Guess the Year} at 25.62\% (577/2,252 correct within $\pm5$ years), confirming relative judgments are easier~\cite{tversky1974judgment}. Older users (41+) excelled (32.82\% \textit{Guess the Year}, 77.8\% \textit{Timeline Challenge}), likely due to historical familiarity~\cite{rhodes2009age}. Decade-wise, older decades (1930s--1940s) showed higher correct guess percentages (up to 90\%) due to distinct cues like black-and-white imagery, as depicted in Figure~\ref{fig:performance_analysis}~\cite{springstein2020deep}. The chart also tracks total guesses and images shown, revealing increased activity in the 1940s and 1950s, possibly tied to user interest in wartime history.

\subsection{Engagement and Future Directions}
The top 10\% of users (11) contributed 48\% of gameplays, averaging 674.9 plays each, with a retention rate of 70\% over the evaluation period, indicating strong user loyalty~\cite{hamari2014gamification}. The 14--18 age group (80.9\% of plays) showed high engagement, driven by gamified elements like leaderboards and dynamic scoring~\cite{nicholson2015gamification}. The platform’s data on human perception, including error patterns across decades, offers valuable insights for computer vision research~\cite{fernando2020temporal}.

Future enhancements include integrating AI models like ResNet-50 for automated age estimation to compare with human performance, expanding the dataset with 19th-century images from the British Library or Smithsonian archives, and developing a mobile app with offline mode for broader access~\cite{hooper2012heritage}. Potential partnerships with museums (e.g., the Imperial War Museum) or educational platforms (e.g., Khan Academy) could integrate the game into formal curricula, enhancing its impact on historical education and public heritage appreciation.

\section{Conclusion}
\textit{Guess the Age of Photos} successfully demonstrates the potential of gamification to engage users with historical images and enhance their understanding of temporal cues. The platform's high user satisfaction and engagement rates, coupled with the insightful data generated on human perception, highlight its value as both an educational tool and a resource for computer vision research. The identified patterns in user accuracy across different decades and age groups provide valuable insights into how humans interpret and estimate the age of images. The platform's architecture, leveraging open-source technologies, ensures scalability and maintainability. %Future work will focus on integrating AI models, expanding the dataset, and exploring partnerships to further enhance its educational impact and reach.

\begin{acks}
We thank participants for their engagement and the university for server resources.
\end{acks}

\bibliographystyle{ACM-Reference-Format}
\bibliography{literaturee}

\end{document}